\PassOptionsToPackage{table}{xcolor}%
\expandafter\def\csname ver@algorithmic.sty\endcsname{}
\documentclass[10pt,letterpaper]{article}
\usepackage{alibaba-report}
\providecommand{\reportaffiliationfont}{}
\title{\raisebox{-0.23\height}{\includegraphics[height=1.6em]{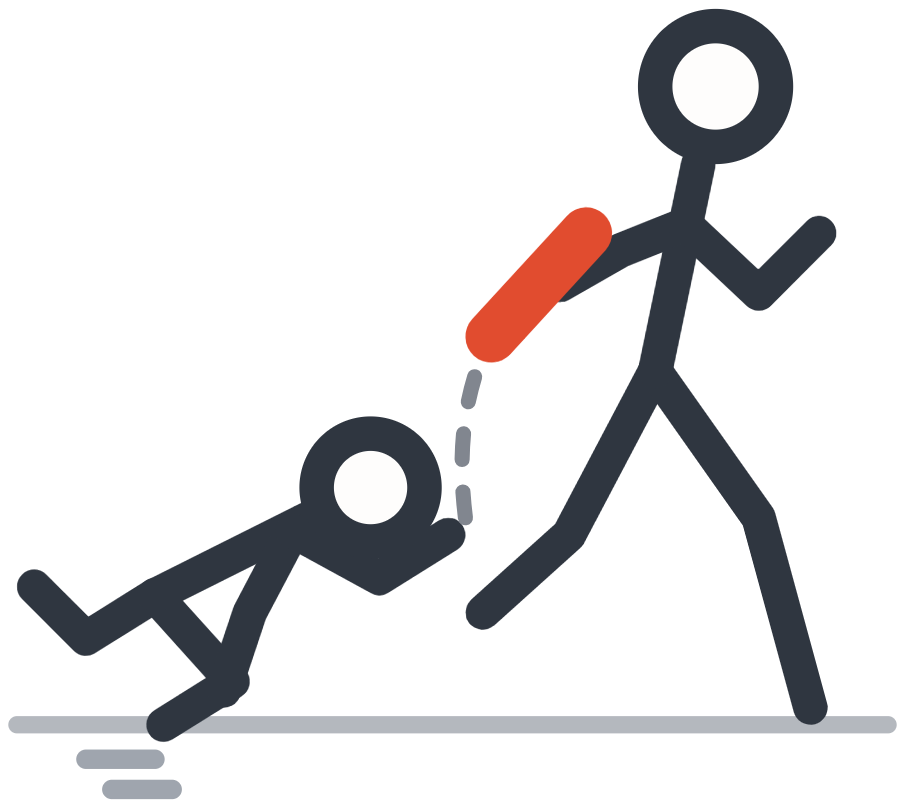}}~Pass the Baton: Trajectory-Relayed On-Policy Distillation}
\author{Haolei Xu\textsuperscript{*,1,2},
  Xiaowen Xu\textsuperscript{*,2},
  Haiwen Hong\textsuperscript{*,$\dagger$,2},
  Zixuan Ni\textsuperscript{1},\\
  Hongxing Li\textsuperscript{1,2},
  Yiwen Qiu\textsuperscript{1},
  Weiming Lu\textsuperscript{$\ddagger$,1},
  Yongliang Shen\textsuperscript{1}\\[6pt]
  {\reportaffiliationfont\mdseries\textsuperscript{1}Zhejiang University,\quad
   \textsuperscript{2}Yuvion Team, Alibaba Group}\\[4pt]
  {\normalfont\textsuperscript{*}Equal contribution,\quad
   \textsuperscript{$\dagger$}Project leader,\quad
   \textsuperscript{$\ddagger$}Corresponding author}\\[3pt]
  {\normalfont\texttt{\{xuhaolei, luwm, syl\}@zju.edu.cn},\quad
   \texttt{honghaiwen.hhw@alibaba-inc.com}}}
\newcommand{\shortpapertitle}{Pass the Baton}
\newcommand{\projecturl}{https://zju-real.github.io/Relay-OPD/}
\newcommand{\codeurl}{https://github.com/zju-real/Relay-OPD}
\renewcommand{\authorcontributions}{}
\drafttrue

\usepackage{array}
\usepackage{makecell}
\usepackage{float}

\usepackage{algpseudocode}
\newcolumntype{L}{>{\raggedright\arraybackslash}X}
\usepackage{newunicodechar}
\newunicodechar{§}{\S}
\newunicodechar{–}{\textendash}
\newunicodechar{—}{\textemdash}
\newunicodechar{→}{\ensuremath{\rightarrow}}
\newunicodechar{∆}{\ensuremath{\Delta}}
\newunicodechar{ć}{\'c}
\definecolor{handoffred}{HTML}{C62828}
\definecolor{deltagreen}{RGB}{0,150,0}
\newtcolorbox{promptbox}[1][]{enhanced,
  breakable,
  colback=black!3,colframe=black!15,boxrule=0.4pt,arc=1mm,
  left=3mm,right=3mm,top=1.5mm,bottom=1.5mm,before skip=8pt,after skip=8pt,
  fonttitle=\sffamily\bfseries\small,coltitle=black!80,colbacktitle=black!7,
  titlerule=0.4pt,toptitle=0.8mm,bottomtitle=0.8mm,
  title=Prompt,
  #1}
\usepackage{needspace}
\renewcommand{\paperlinks}{\par\vspace{3pt}\begingroup
  \sffamily\fontsize{7.5}{10}\selectfont
  \noindent\makebox[\linewidth][c]{\scalebox{.9}{\paperlinkentry{\projecturl}{Project}}\hspace{24pt}%
  \scalebox{.9}{\paperlinkentry{\codeurl}{Code}}}
  \par\endgroup}

\begin{document}
\maketitle
\begin{abstract}
On-policy distillation (OPD) grounds token-level supervision in the student's own trajectory, yet suffers from prefix failure: once the student commits to a wrong reasoning direction, all subsequent generation builds on this deviation, producing misdirected continuations that elicit unreliable supervision and waste compute. We identify a teacher--student continuation asymmetry on failed prefixes, where the teacher tends to redirect while the student continues along the original direction, and convert it into a label-free handoff trigger in Relay On-Policy Distillation (Relay-OPD). During training, Relay-OPD constructs relay trajectories by letting the teacher briefly take over at detected trigger points to produce a teacher leg, after which the student resumes and is optimized on the resulting trajectory. A limited relay budget concentrates intervention on critical early positions while limiting departure from the student policy. With a Qwen3-4B-Instruct-2507 teacher and Qwen3-0.6B/1.7B-Non-Thinking students on eight mathematical reasoning benchmarks, Relay-OPD achieves the best or second-best results on every benchmark, outperforming standard OPD by +5.73\% and the strongest baseline FastOPD by +1.49\% on average for 1.7B, with consistent gains at 0.6B. Training trajectory length is reduced by over 50\%.

\paperlinks
\end{abstract}
\begin{figure}[H]
  \centering
  \includegraphics[width=\textwidth]{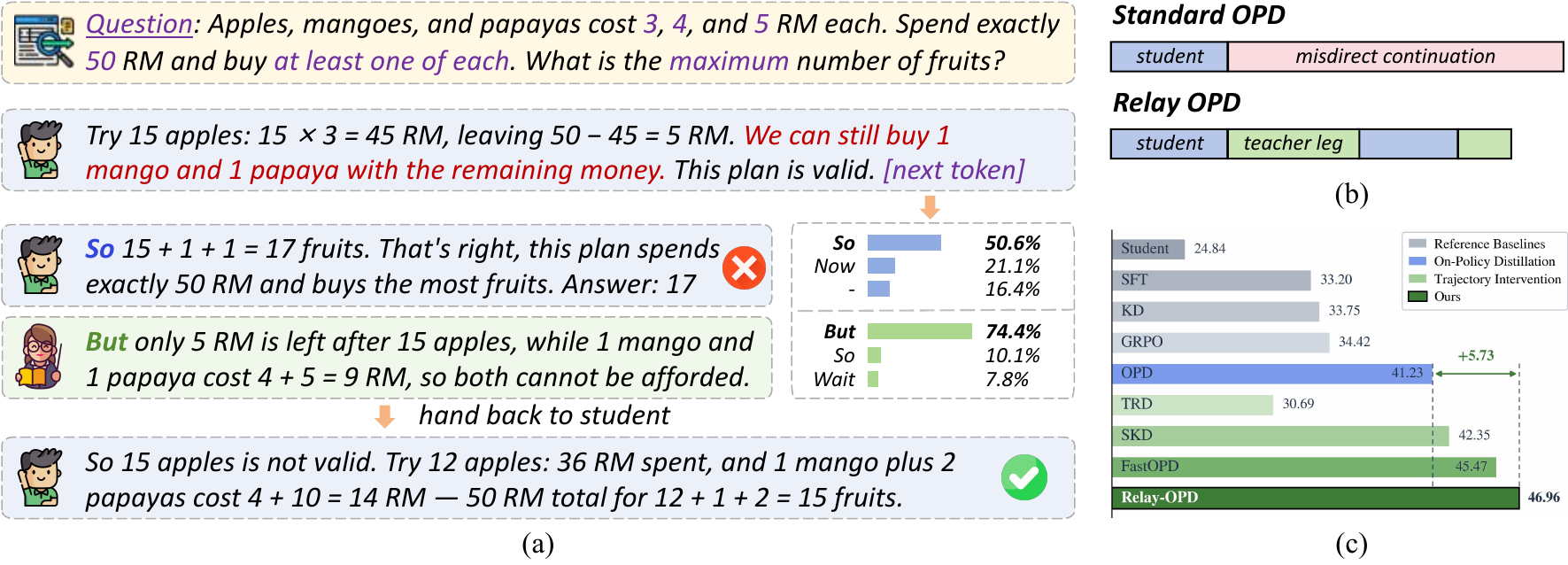}
  \caption{\textbf{(a)} A Relay-OPD case: at the detected handoff trigger, the teacher prefers the reflection token (\texttt{But}, 74.4\%) while the student would continue along the current direction (\texttt{So}, 50.6\%); a brief teacher leg corrects the reasoning and the student resumes to the correct answer. \textbf{(b)} Difference between OPD and Relay-OPD. \textbf{(c)} Overall performance.}
  \label{fig:teaser}
\end{figure}

\clearpage
\section{Introduction}\label{sec:intro}

Effectively transferring capabilities from strong models to weaker ones has become a central challenge in large language model post-training \citep{yang2025qwen3,xu2026deepseek,zeng2026glm,xiao2026mimo}. Unlike supervised fine-tuning and offline knowledge distillation \citep{hinton2015distilling,kim2016sequence}, which both rely on teacher-generated data, on-policy distillation (OPD) \citep{agarwal2024policy,lu2025onpolicydistillation} lets the student generate trajectories from its own policy. The teacher then supplies dense token-level guidance on the prefixes the student actually visits. Because supervision is always grounded in the student's own state distribution, OPD effectively mitigates the train--inference distribution shift, and has shown clear gains in strong-to-weak distillation.

Yet the student's own trajectories inevitably include its failures.
In long-chain reasoning \citep{jaech2024openai,guo2025deepseek}, this leads to \textbf{prefix failure} \citep{li2026rethinking,fu2026revisiting,xie2026position}: once the student commits to a wrong direction early on, all subsequent generation builds on this deviation (Figure~\ref{fig:teaser}(b)). The resulting long misdirected continuations elicit unreliable and potentially harmful supervision, and they waste substantial training compute.

Prior work addresses prefix failure from several angles, each with a structural limitation. Fixed-length truncation (ESR, FastOPD) \citep{ziheng2026less,zhang2026fast} cuts the rollout at a rigid position regardless of where the reasoning actually fails. Offline rewriting (TRD) \citep{jiang2026trajectory} repairs trajectories only after the rollout completes, and the repairs often leave visible artifacts. Token-level mixing (SKD) \citep{xu2025speculative} switches between teacher and student on generic distributional disagreement, not on an explicit signal that the reasoning direction has failed. What is missing is a mechanism that acts \emph{online}, correcting failure as it emerges, and that decides \emph{where} to intervene from the reasoning state itself.

On a failed student prefix, the teacher and student diverge in how they continue (Figure~\ref{fig:teaser}(a); the complete case appears in Appendix~\ref{app:takeover-case}): the teacher tends to stop, re-examine the reasoning, and redirect, whereas the student tends to press on in the same wrong direction. This divergence is observable during generation, without any label, verifier, or reward model, and it marks where the student's reasoning has gone wrong. We call such a position a \textbf{handoff trigger}.
To find out how a trigger should be handled, we run trajectory intervention experiments: at each trigger the teacher takes over briefly, and we vary the duration and timing of the intervention.

\begin{figure}[t]
  \centering
  \begin{subfigure}[b]{0.565\textwidth}
    \centering
    \includegraphics[width=\linewidth]{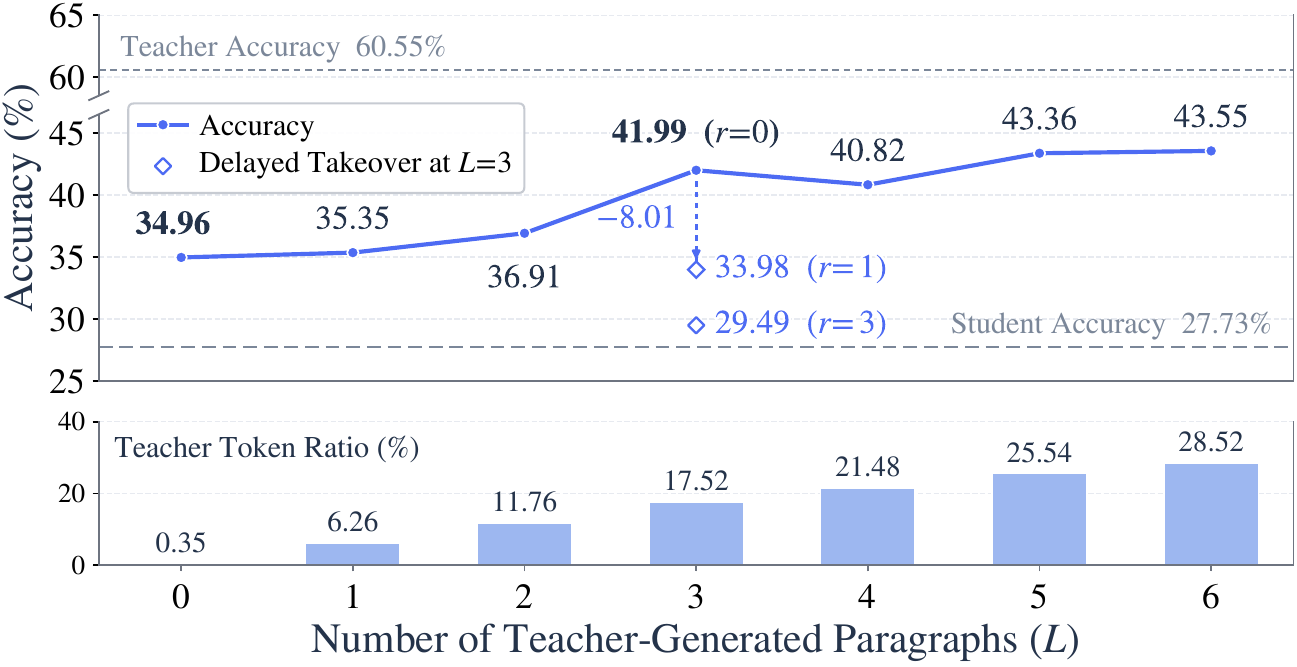}
    \caption{}
    \label{fig:motivation-intervention}
  \end{subfigure}\hfill
  \begin{subfigure}[b]{0.425\textwidth}
    \centering
    \includegraphics[width=\linewidth]{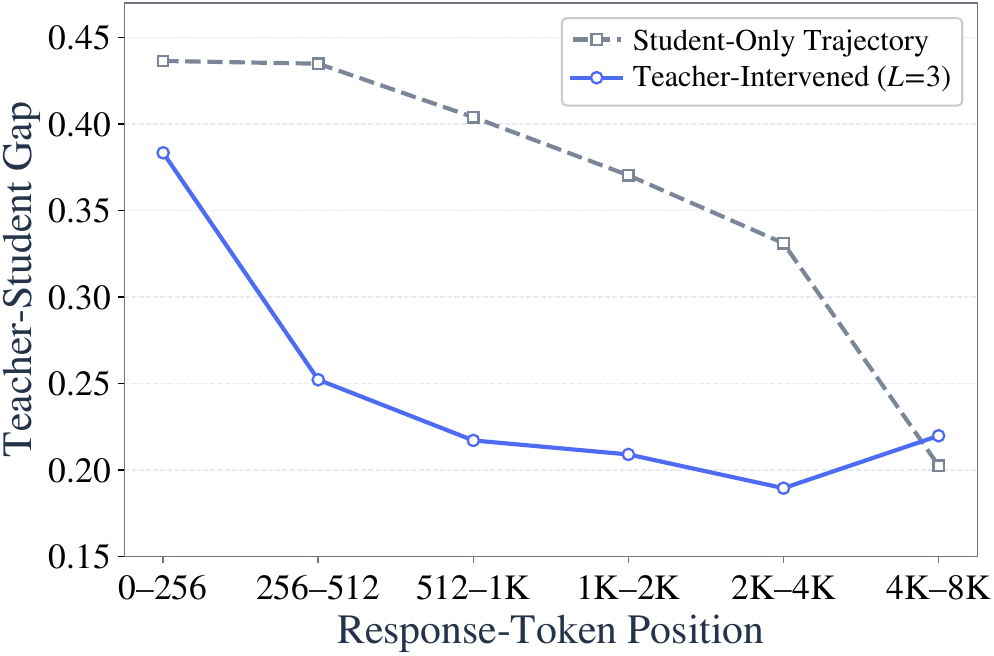}
    \caption{}
    \label{fig:absolute-position-gap}
  \end{subfigure}
  \caption{\textbf{Preliminary Trajectory Intervention Experiments.} Qwen3-4B-Instruct-2507 (teacher) and Qwen3-1.7B-Non-Thinking (student) on 128 DAPO-Math-17K English samples; the handoff criterion matches the main experiments. \textbf{(a)} Accuracy (mean@4) and teacher token ratio as the number of teacher-generated paragraphs $L$ varies; a periodic delay of $r$ skips $r$ consecutive valid triggers before the next takeover. \textbf{(b)} Teacher--student gap by response-token position under the $L{=}3$ intervention, computed as the absolute value of the mean token-level advantage (defined in \S\ref{subsec:opd}) within each position interval.}
  \label{fig:motivation}
\end{figure}

Two properties of this intervention stand out (Figure~\ref{fig:motivation}). First, correction can be remarkably \emph{local}: replacing only the single reflection token at each trigger, so that teacher tokens are merely $0.35\%$ of all generated tokens, already lifts accuracy from 27.73 to 34.96 (+7.23\%; Figure~\ref{fig:motivation-intervention}), and teacher takeover likewise reduces the teacher--student gap (Figure~\ref{fig:absolute-position-gap}). Second, the intervention's value is \emph{front-loaded}: holding the intervention length fixed but shifting it to later triggers instead of the earliest ones drops accuracy from 41.99 to 33.98 and then 29.49. The teacher--student gap explains why timing dominates. It narrows as generation proceeds, whether the student runs alone or the teacher finishes the trajectory. As the prefix grows, the teacher is pulled along by the student's context, so a late takeover cannot redirect it.

A clear design principle emerges: prefix failure can be addressed through early, local teacher intervention at detected failure points, before the wrong direction becomes entrenched in the trajectory.
This leaves two questions for a training method: \emph{when} should the teacher take over, and how do we keep these takeovers from pulling the trajectory too far from the student's own policy?

We propose \textbf{Relay On-Policy Distillation (Relay-OPD)}, which interleaves student generation with brief teacher takeovers at the points where reasoning first goes wrong. As the student rolls out its trajectory, Relay-OPD monitors each prefix for a handoff trigger: a state where the teacher would redirect the reasoning while the student would continue on its current course. When a trigger fires, the teacher generates a short \emph{teacher leg} that steers the reasoning back and returns control to the student (Figure~\ref{fig:teaser}(b)). A \emph{relay budget} caps the number and length of these legs, concentrating correction at the early positions where prefix failure originates while keeping the trajectory close to the student's own policy. The student is then distilled on the resulting relay trajectory, exactly as in standard on-policy distillation.

Our contributions are as follows: (1)~We identify a \textbf{teacher--student continuation asymmetry} on failed reasoning prefixes and show, through trajectory intervention experiments, that correcting prefix failure needs only early and local teacher intervention, with teacher tokens as low as $0.35\%$. (2)~We propose \textbf{Relay-OPD}, which turns this asymmetry into a label-free handoff trigger and a budgeted relay rollout, and unifies student and teacher generation in a single speculative-decoding engine. (3)~Across eight math reasoning benchmarks and two student scales, Relay-OPD outperforms standard OPD by $+5.73\%$ and FastOPD by $+1.49\%$ for the 1.7B student, while cutting average training trajectory length by over $50\%$.

\section{Related Work}\label{sec:related}

\subsection{On-Policy Distillation}\label{subsec:online-distillation}

On-policy distillation samples trajectories from the student, with the teacher scoring each token on the prefixes the student actually visits \citep{agarwal2024policy,gu2024minillm}; compared with offline training on teacher-generated data \citep{hinton2015distilling,kim2016sequence}, it grounds supervision in the student's own state distribution, yields faster and stronger transfer, and is now standard in post-training pipelines \citep{yang2025qwen3,lu2025onpolicydistillation}. Recent extensions cover distillation across model families \citep{patino2025unlocking}, black-box teachers without logit access \citep{ye2025black}, and self-distillation of a model's own in-context or privileged knowledge \citep{yang2025distilling,hubotter2026reinforcement,shenfeld2026self,zhao2026self,penaloza2026privileged}.\par

\subsection{Prefix Failure}\label{subsec:prefix-failure}

Grounding supervision entirely in student trajectories also imports the student's failures. Distillation gains shrink when teacher and student reasoning patterns are incompatible \citep{li2026rethinking}, and as student prefixes drift from teacher-supported states, subsequent supervision grows unreliable \citep{fu2026revisiting,xie2026position}. These effects are most pronounced in long-chain reasoning, where early directional deviations compound autoregressively into extended misdirected continuations. Existing remedies intervene on the trajectory itself: ESR and FastOPD \citep{ziheng2026less,zhang2026fast} truncate rollouts at a fixed length to discard late, low-value supervision, TRD \citep{jiang2026trajectory} rewrites student trajectories offline via the teacher, and SKD \citep{xu2025speculative} mixes teacher and student generation token by token according to distributional agreement.\par

\section{Method}\label{sec:method}

This section first reviews standard on-policy distillation (\S\ref{subsec:opd}), then introduces the handoff trigger, relay trajectory construction, and optimization objective of Relay-OPD (\S\ref{subsec:relay-opd}), and finally describes its efficient implementation (\S\ref{subsec:implementation}).\par

\subsection{On-Policy Distillation}\label{subsec:opd}

Let $x\sim\mathcal D$ denote a training prompt, and let $\pi_\theta$, $\pi_{\bar\theta}$, and $\pi_T$ denote the current student policy, old student policy, and teacher policy. At each training iteration, standard OPD first generates an on-policy trajectory using the old student policy:\par

\begin{equation}
y=(y_1,\ldots,y_N)\sim\pi_{\bar\theta}(\cdot\mid x).
\end{equation}

Let $h_t=(x,y_{<t})$ denote the student prefix at position $t$. Standard OPD typically employs the reverse KL divergence. Taking a single-sample estimate \citep{lu2025onpolicydistillation} at student-sampled token $y_t$:\par

\begin{equation}
\widehat D_{\mathrm{RKL}}^{(t)}
=\log\pi_{\bar\theta}(y_t\mid h_t)
-\log\pi_T(y_t\mid h_t).
\end{equation}

Its negation serves as the advantage:\par

\begin{equation}
A_t^{\mathrm{OPD}}
=-\widehat D_{\mathrm{RKL}}^{(t)}
=\log\pi_T(y_t\mid h_t)
-\log\pi_{\bar\theta}(y_t\mid h_t).
\end{equation}

A positive advantage encourages the student to increase the probability of the sampled token; a negative one decreases it. Since training is entirely grounded in student-visited prefixes, once prefix failure occurs early in generation, the resulting long misdirected continuation is still included in training despite receiving increasingly unreliable supervision.\par

\subsection{Relay-OPD}\label{subsec:relay-opd}

\begin{figure}[t]
  \centering
  \includegraphics[width=\textwidth]{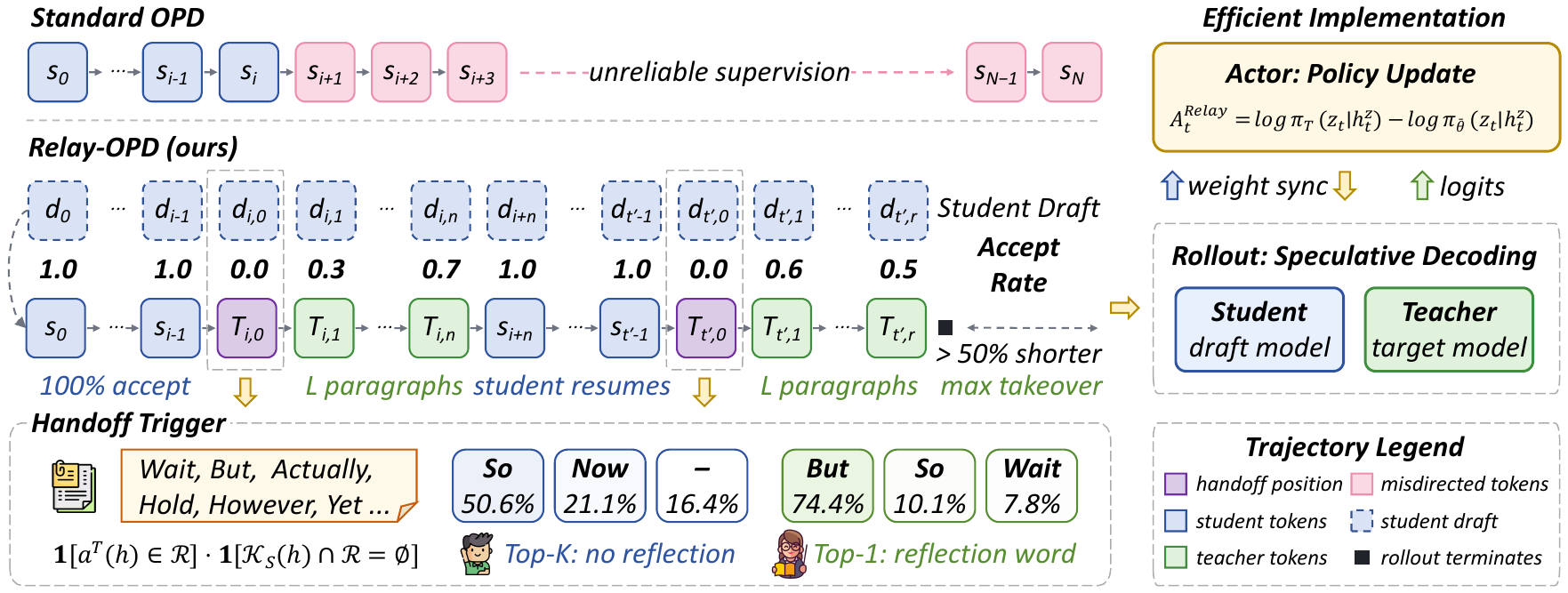}
  \caption{\textbf{Overview of Relay-OPD.} Unlike standard OPD (top), which trains on student-only trajectories that continue misdirected after prefix failure, Relay-OPD constructs relay trajectories (middle): when the handoff trigger (bottom) detects that the teacher would redirect the reasoning while the student would not, the teacher briefly takes over before the student resumes. The entire rollout runs in a single speculative decoding engine (right; \S\ref{subsec:implementation}).}
  \label{fig:method}
\end{figure}

Relay-OPD introduces state-driven teacher takeover into the student trajectories of standard OPD. The method first detects handoff triggers based on teacher--student continuation tendencies on the current prefix, then constructs a relay trajectory comprising student legs and teacher legs, and performs distillation on the resulting trajectory. Figure~\ref{fig:method} provides an overview.\par

\textbf{Handoff trigger.} Identifying prefix failure online requires a signal that depends on neither external verifiers nor process labels. Rather than relying on generic teacher--student distributional differences, we focus on divergences in reasoning direction: takeover is triggered when the teacher tends to redirect while the student tends to continue along the current direction (Figure~\ref{fig:method}, bottom).\par

Let $\mathcal R\subseteq\mathcal V$ denote a set of reflection tokens used for redirecting reasoning \citep{guo2025deepseek,muennighoff2025s1}, such as Wait, But, and However, together with their case and leading-space variants; the complete list is provided in Appendix~\ref{app:trigger-words}. Given a student prefix $h$, define the teacher's most probable next token as\par

\begin{equation}
a^T(h)=\mathop{\arg\max}_{v\in\mathcal V}\pi_T(v\mid h),
\end{equation}

and the student's top-$K$ support set on this prefix as\par

\begin{equation}
\mathcal K_S(h)
=\operatorname{TopK}_K\!\left(\pi_{\bar\theta}(\cdot\mid h)\right).
\end{equation}

When the teacher's preferred token belongs to $\mathcal R$ while the student's top-$K$ support set contains no token from $\mathcal R$, we define the handoff criterion:\par

\begin{equation}
\phi(h)
=\mathbf 1[a^T(h)\in\mathcal R]
\cdot
\mathbf 1[\mathcal K_S(h)\cap\mathcal R=\varnothing].
\end{equation}

$K$ controls the divergence threshold required for triggering and thus the sensitivity of the criterion.\par

\textbf{Relay trajectory construction.} Two findings from the preliminary experiments shape the design. First, the benefit of extending teacher takeover saturates (Figure~\ref{fig:motivation-intervention}): from $L=3$ to $L=6$, accuracy plateaus around 41--44 despite the teacher token ratio rising from 17.52\% to 28.52\%, motivating a limited teacher leg. Second, valuable supervision concentrates early in the trajectory (Figure~\ref{fig:absolute-position-gap}), motivating a relay budget that caps the total number of takeovers and focuses intervention on earlier positions.\par

The relay budget $(M,L)$ specifies the maximum number of teacher takeovers $M$ and the number of additional paragraphs $L$ generated after the reflection token in each teacher leg, with paragraphs delimited by \texttt{\textbackslash{}n\textbackslash{}n}. We measure teacher legs in paragraphs rather than a fixed number of tokens so that each leg ends at a structurally complete reasoning unit instead of breaking off mid-thought; in our setting a paragraph contains 23.2 tokens on average. The student generates the student leg autoregressively according to $\pi_{\bar\theta}$, computing $\phi(h)$ at each position. When $\phi(h)=1$ and the takeover count has not reached $M$, the teacher takes over: $a^T(h)$ from the trigger criterion serves as the starting token of the teacher leg, and generation continues for $L$ paragraphs. When $L=0$, the segment consists solely of the reflection token. After the teacher leg, if budget remains, the student resumes generation from the extended prefix; when the $M$-th teacher leg ends, the current rollout terminates. This yields the relay trajectory $z=(z_1,\ldots,z_N)$.

Unlike fixed-length truncation, both the intervention and termination positions of Relay-OPD are determined by the current reasoning state.\par

\textbf{Optimization objective.} Figure~\ref{fig:motivation-intervention} shows that although teacher takeover significantly reduces prefix failure, even at $L=6$ with a teacher token ratio of 28.52\%, accuracy reaches only 43.55 compared with 60.55 for the teacher generating independently. We therefore posit that the training signal the teacher provides on relay trajectories differs from its ideal supervision on its own trajectories. Accordingly, we adopt a reverse-KL-style single-sample objective that optimizes directly on the observed tokens in the relay trajectory, enabling the student to selectively absorb the teacher's corrective signals rather than fully fitting the teacher distribution via forward KL.\par

At position $t$, let $h_t^z=(x,z_{<t})$ denote the relay prefix. The advantage for the actually generated token $z_t$ is\par

\begin{equation}
A_t^{\mathrm{Relay}}
=\log\pi_T(z_t\mid h_t^z)
-\log\pi_{\bar\theta}(z_t\mid h_t^z).
\end{equation}

The update ratio of the current student relative to the old student is\par

\begin{equation}
\rho_t(\theta)
=\frac{\pi_\theta(z_t\mid h_t^z)}
{\pi_{\bar\theta}(z_t\mid h_t^z)}.
\end{equation}

Relay-OPD optimizes:\par

\begin{equation}
\mathcal L_{\mathrm{Relay}}
=-\mathbb E_{x,\,z}
\left[
\frac{1}{N}\sum_{t=1}^{N}
\min\left(
\rho_tA_t^{\mathrm{Relay}},
\operatorname{clip}(\rho_t,1-\epsilon,1+\epsilon)A_t^{\mathrm{Relay}}
\right)
\right].
\end{equation}

This objective uses the actually generated tokens $z_t$ across the entire relay trajectory. Teacher legs both provide corrected context for subsequent student legs and directly participate in optimization through their generated tokens.\par

When $\phi(h)\equiv0$, the relay trajectory degenerates to a standard student trajectory; when $L=0$, the teacher leg contains only the replacement token at the handoff trigger position, corresponding to minimal prefix correction. The complete algorithm is given in Appendix~\ref{app:relay-algorithm}.\par

\subsection{Efficient Implementation}\label{subsec:implementation}

Relay-OPD requires continuously determining whether to trigger teacher takeover during generation and alternating between student and teacher legs. Maintaining two independent generation pipelines would necessitate repeated coordination between teacher and student models for takeover timing, plus switching generation engines at takeover and recovery points, introducing substantial scheduling and communication overhead. We instead unify the entire Relay-OPD generation process within a single speculative decoding engine \citep{leviathan2023fast,chen2023accelerating}: the student serves as draft model and the teacher as target model (Figure~\ref{fig:method}, right).\par

\textbf{Relay as a state-switched decoding process.} The engine produces the relay trajectory $z$ of \S\ref{subsec:relay-opd} one position at a time: at each position $t$, it first determines a decoding state $s_t\in\{\mathsf{S},\mathsf{T},\bot\}$ from the current prefix $h_t^z=(x,z_{<t})$, and then emits $z_t$ in a student leg ($\mathsf{S}$) or a teacher leg ($\mathsf{T}$), or terminates ($\bot$). Let $j_t$ denote the number of teacher takeovers and $\ell_t$ the number of completed paragraphs (delimiter \texttt{\textbackslash n\textbackslash n}) in the current teacher leg, both up to position $t$. Starting from $s_1=\mathsf{S}$, the state for the next position is determined by
\begin{equation}
s_{t+1}=
\begin{cases}
\mathsf{T}, & \text{if } s_t=\mathsf{S},\ \phi(h_{t+1}^z)=1,\ \text{and } j_t<M,\\[2pt]
\mathsf{S}, & \text{if } s_t=\mathsf{T},\ \ell_t=L,\ \text{and } j_t<M,\\[2pt]
\bot, & \text{if } s_t=\mathsf{T},\ \ell_t=L,\ \text{and } j_t=M,\\[2pt]
s_t, & \text{otherwise},
\end{cases}
\label{eq:state-transition}
\end{equation}
and additionally enters the absorbing state $\bot$ whenever $z_t$ is the end-of-sequence token or the length limit is reached. The first case realizes the handoff trigger: a trigger detected on the prefix $h_{t+1}^z$ switches the state to $\mathsf{T}$, so the teacher takes over at position $t+1$. When $L=0$, the exit condition $\ell_t=L$ already holds at the leg-initial position, so the teacher leg consists of exactly the reflection token, recovering the minimal-correction case of \S\ref{subsec:relay-opd}.\par

\textbf{Unified token generation.} Every position is generated by speculative decoding against a state-dependent target policy: $\pi^{\mathrm{tgt}}_t=\pi_{\bar\theta}$ when $s_t=\mathsf{S}$, and $\pi^{\mathrm{tgt}}_t=\pi_T$ when $s_t=\mathsf{T}$. In student legs every draft is accepted and generation reduces to ordinary student decoding, while teacher legs perform standard speculative decoding against the teacher. The single exception is the leg-initial position of each teacher leg, where the engine directly emits the trigger token $z_t=a^T(h_t^z)$ without verification. At every other position, the student drafts $a_t^S\sim\pi_{\bar\theta}(\cdot\mid h_t^z)$, which is accepted with probability
\begin{equation}
\alpha_t
=\min\!\left(1,\;
\frac{\pi^{\mathrm{tgt}}_t(a_t^S\mid h_t^z)}
{\pi_{\bar\theta}(a_t^S\mid h_t^z)}\right);
\label{eq:acceptance}
\end{equation}
if an i.i.d.\ draw $u_t\sim\mathcal U(0,1)$ exceeds $\alpha_t$, the draft is rejected and $z_t$ is instead sampled from the residual distribution
\begin{equation}
q_t(v)=
\frac{\bigl[\pi^{\mathrm{tgt}}_t(v\mid h_t^z)-\pi_{\bar\theta}(v\mid h_t^z)\bigr]_+}
{\sum_{w\in\mathcal V}
\bigl[\pi^{\mathrm{tgt}}_t(w\mid h_t^z)-\pi_{\bar\theta}(w\mid h_t^z)\bigr]_+},
\qquad [x]_+=\max(x,0).
\label{eq:residual}
\end{equation}
In student legs, substituting $\pi^{\mathrm{tgt}}_t=\pi_{\bar\theta}$ into Eq.~\eqref{eq:acceptance} gives $\alpha_t\equiv1$, so drafts are accepted unconditionally; in teacher legs, Eqs.~\eqref{eq:acceptance}--\eqref{eq:residual} are exactly standard speculative rejection sampling against $\pi_T$.\par

\textbf{Exactness and efficiency.} By the correctness of speculative sampling \citep{leviathan2023fast}, every verified position satisfies $z_t\sim\pi^{\mathrm{tgt}}_t(\cdot\mid h_t^z)$. Hence, conditioned on the prefix and the deterministic leg-initial token, teacher legs are distributionally identical to the teacher continuing generation directly from the extended prefix, and student legs to ordinary student sampling: the single-engine implementation reproduces the two-model relay process of \S\ref{subsec:relay-opd} exactly, without switching between two generation pipelines. Efficiency-wise, teacher legs batch-verify student drafts instead of decoding serially, and the teacher logits computed during verification simultaneously provide $a^T(h_t^z)$ and the trigger criterion $\phi(h_t^z)$ at no additional cost. In practice, Eq.~\eqref{eq:state-transition} is applied during block verification: draft tokens following the first transition point within a block are discarded, so the per-token semantics above are preserved exactly.\par

After each parameter update, the latest student weights are synchronized to the draft model; teacher parameters remain frozen throughout training.\par

\section{Experiments}\label{sec:experiments}

\subsection{Experimental Setup}\label{subsec:setup}

\textbf{Models, data, and benchmarks.} We use Qwen3-4B-Instruct-2507 \citep{yang2025qwen3} as teacher and Qwen3-0.6B-Non-Thinking and Qwen3-1.7B-Non-Thinking as students. Training data is the English subset of DAPO-Math-17K \citep{yu2026dapo}. Evaluation covers eight mathematical reasoning benchmarks: AIME 2024 \citep{aime2024}, AIME 2025, AIME 2026, MATH500 \citep{hendrycks2021measuring,lightman2024let}, AMC 2023 \citep{amc2023}, OlympiadBench \citep{he2024olympiadbench}, HMMT February 2026 \citep{hmmt25}, and HMMT November 2025. At evaluation time, we set temperature to 1.0 and top-$p$ to 1.0, with a maximum generation length of 32,768 tokens. Each problem in AIME, AMC, and HMMT is sampled 32 times; each problem in MATH500 and OlympiadBench is sampled 4 times. We report mean accuracy.\par

\textbf{Implementation details.} All methods are implemented on verl \citep{sheng2025hybridflow} and vLLM 0.21.0 \citep{kwon2023efficient}, trained on 8 H100 GPUs. Online distillation methods train for 1 epoch with a maximum response length of 16,384 tokens. Trajectory sampling uses temperature 1.0 and top-$p=1.0$. Relay-OPD sets $K=5$ and $(M,L)=(2,3)$, without using any external verifier, process supervision labels, or answer correctness labels. Complete hyperparameters are provided in Appendix~\ref{app:hyperparameters}, and the training and inference prompt template in Appendix~\ref{app:training-prompt}.\par

\textbf{Baselines.} We compare Relay-OPD against three categories of methods.\par

\begin{enumerate}[leftmargin=*,itemsep=0.25ex,topsep=0.5ex]
\item \textbf{Reference baselines.} The untrained initial student, supervised fine-tuning (SFT), token-level knowledge distillation (KD) \citep{hinton2015distilling}, and the outcome-reward reinforcement learning method GRPO \citep{shao2024deepseekmath}.
\item \textbf{On-policy distillation.} Standard OPD \citep{lu2025onpolicydistillation}, which performs token-level distillation on student-generated trajectories.
\item \textbf{Trajectory intervention methods.} TRD \citep{jiang2026trajectory}, which rewrites student trajectories offline via the teacher; FastOPD \citep{ziheng2026less,zhang2026fast}, which shortens the rollout budget; and SKD \citep{xu2025speculative}, which mixes teacher and student generation through speculative decoding. For FastOPD, we evaluate fixed truncation lengths in $\{1024,2048,4096,8192\}$; Table~\ref{tab:main-results} reports the best configuration at 4,096 tokens, with complete results in Appendix~\ref{app:fastopd-sweep}.
\end{enumerate}

Implementation details for each method are provided in Appendix~\ref{app:algorithm-baselines}, with a structured overview in Table~\ref{tab:method-comparison}.\par

\subsection{Main Results}\label{subsec:main-results}

\begin{table}[t]
\centering
\caption{\textbf{Main Experimental Results.} Mean accuracy is reported; bold and underline denote the best and second-best results for each student model. Subscripts in the Avg column indicate the training step of the best checkpoint. Train Len denotes the average rollout response length from the start of training to the best checkpoint.}
\label{tab:main-results}
\begingroup
\small
\newcommand{\dv}[1]{\textcolor{deltagreen}{#1}}
\renewcommand{\arraystretch}{1.1}
\setlength{\tabcolsep}{3pt}
\resizebox{\linewidth}{!}{%
\begin{tabular}{lcccccccccc}
\toprule
\textbf{Method} & \textbf{AIME24} & \textbf{AIME25} & \textbf{AIME26} & \textbf{MATH} & \textbf{AMC23} & \textbf{Olymp.} & \textbf{\makecell{HMMT\\Feb26}} & \textbf{\makecell{HMMT\\Nov25}} & \textbf{Avg} & \textbf{\makecell{Train\\Len}} \\
\midrule
Teacher & 60.42 & 46.04 & 52.19 & 94.20 & 93.83 & 70.62 & 31.25 & 41.88 & 61.30 & — \\
\midrule
\multicolumn{11}{c}{\textit{Student: Qwen3-0.6B-Non-Thinking}} \\
\midrule
Student & 1.77 & 2.40 & 0.73 & 44.10 & 24.45 & 16.36 & 0.76 & 3.85 & 11.80 & — \\
SFT & 4.90 & 7.60 & 4.06 & 59.45 & 34.92 & 26.74 & 1.89 & 3.02 & 17.82$_{\text{@110}}$ & 4262 \\
KD & 4.17 & 7.19 & 4.79 & 57.75 & 35.23 & 27.15 & 2.94 & 2.71 & 17.74$_{\text{@110}}$ & 4262 \\
GRPO & 8.23 & 15.21 & 10.00 & 68.60 & 46.56 & 35.42 & 7.86 & 6.04 & 24.74$_{\text{@110}}$ & 3379 \\
OPD & 13.44 & 17.92 & 11.98 & 75.30 & 51.25 & 41.32 & 7.39 & 5.62 & 28.03$_{\text{@110}}$ & 6900 \\
TRD & 4.58 & 8.54 & 3.96 & 57.60 & 36.33 & 26.93 & 3.79 & 2.81 & 18.07$_{\text{@20}}$ & 3275 \\
FastOPD & \underline{15.83} & \underline{20.10} & \underline{13.54} & \underline{75.30} & \underline{53.67} & \underline{44.81} & \textbf{11.55} & \underline{8.54} & \underline{30.42}$_{\text{@100}}$ & 3302 \\
SKD & 8.44 & 16.98 & 9.79 & 66.65 & 43.67 & 35.76 & 8.14 & 5.62 & 24.38$_{\text{@20}}$ & 5800 \\
\rowcolor{cyan!10}
\textbf{Relay-OPD} & \textbf{15.94} & \textbf{20.94} & \textbf{14.06} & \textbf{76.80} & \textbf{55.55} & \textbf{45.03} & \underline{11.17} & \textbf{8.85} & \textbf{31.04}$_{\text{@75}}$ & \textbf{2490} \\
\rowcolor{cyan!10}
\dv{$\Delta$ vs OPD} & \dv{+2.50} & \dv{+3.02} & \dv{+2.08} & \dv{+1.50} & \dv{+4.30} & \dv{+3.71} & \dv{+3.78} & \dv{+3.23} & \dv{+3.01} & \dv{-63.9\%} \\
\midrule
\multicolumn{11}{c}{\textit{Student: Qwen3-1.7B-Non-Thinking}} \\
\midrule
Student & 12.60 & 9.58 & 7.40 & 71.95 & 47.89 & 38.54 & 6.34 & 4.38 & 24.84 & — \\
SFT & 23.33 & 19.48 & 16.15 & 81.40 & 59.45 & 46.62 & 12.59 & 6.56 & 33.20$_{\text{@110}}$ & 4262 \\
KD & 23.54 & 21.15 & 15.31 & 81.45 & 60.23 & 48.07 & 12.78 & 7.50 & 33.75$_{\text{@110}}$ & 4262 \\
GRPO & 24.58 & 22.08 & 15.62 & 80.35 & 60.16 & 48.74 & 14.49 & 9.38 & 34.42$_{\text{@105}}$ & 2558 \\
OPD & 35.83 & 25.52 & 23.33 & 85.70 & 70.08 & 55.27 & 20.08 & 14.06 & 41.23$_{\text{@55}}$ & 4658 \\
TRD & 19.27 & 19.69 & 12.71 & 77.70 & 55.47 & 44.18 & 11.93 & 4.58 & 30.69$_{\text{@40}}$ & 2785 \\
FastOPD & \underline{42.29} & 30.42 & 26.35 & \underline{87.95} & \underline{74.30} & \underline{58.16} & \underline{23.58} & \textbf{20.73} & \underline{45.47}$_{\text{@45}}$ & 2709 \\
SKD & 33.12 & \underline{30.73} & \underline{28.85} & 87.35 & 72.42 & 54.41 & 20.08 & 11.88 & 42.35$_{\text{@35}}$ & 4753 \\
\rowcolor{cyan!10}
\textbf{Relay-OPD} & \textbf{42.71} & \textbf{32.81} & \textbf{30.52} & \textbf{89.50} & \textbf{76.88} & \textbf{58.79} & \textbf{24.72} & \underline{19.79} & \textbf{46.96}$_{\text{@35}}$ & \textbf{2296} \\
\rowcolor{cyan!10}
\dv{$\Delta$ vs OPD} & \dv{+6.88} & \dv{+7.29} & \dv{+7.19} & \dv{+3.80} & \dv{+6.80} & \dv{+3.52} & \dv{+4.64} & \dv{+5.73} & \dv{+5.73} & \dv{-50.7\%} \\
\bottomrule
\end{tabular}%
}
\endgroup
\end{table}

\begin{figure}[tb]
  \centering
  \begin{minipage}[t]{0.60\textwidth}
    \centering
    \includegraphics[width=\linewidth]{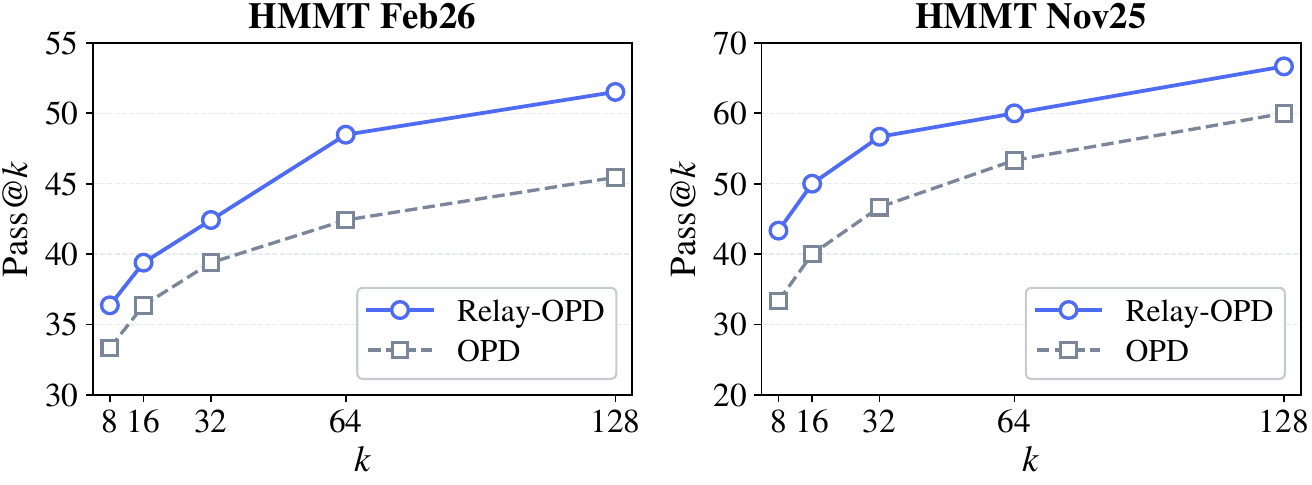}
    \caption{Pass@$k$ performance of Relay-OPD and OPD on HMMT Feb26 and HMMT Nov25.}
    \label{fig:hmmt-passk}
  \end{minipage}\hfill
  \begin{minipage}[t]{0.37\textwidth}
    \centering
    \includegraphics[width=\linewidth]{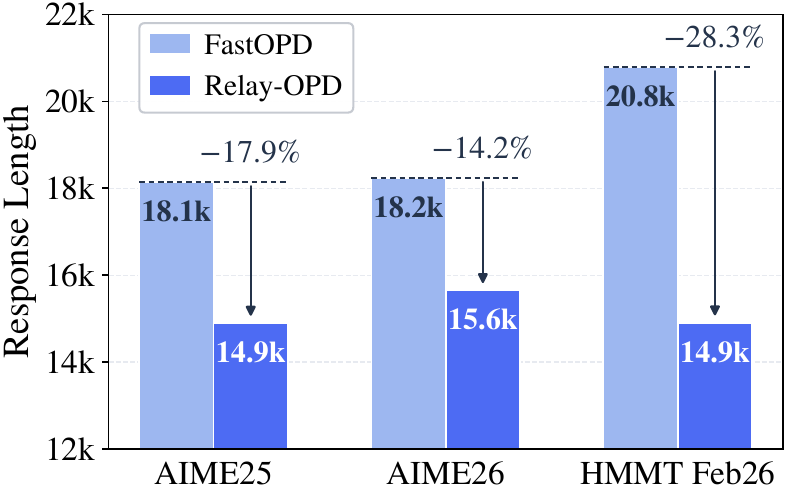}
    \caption{Inference response length of Relay-OPD vs.\ FastOPD.}
    \label{fig:inference-length}
  \end{minipage}
\end{figure}

\textbf{Overall performance.} As summarized in Table~\ref{tab:main-results}, Relay-OPD achieves the best or second-best results across all eight benchmarks for both student models. For Qwen3-1.7B-Non-Thinking, it attains an average accuracy of 46.96, outperforming standard OPD by +5.73\% and the strongest trajectory intervention baseline FastOPD by +1.49\%; AIME 2025 and AIME 2026 improve over OPD by +7.29\% and +7.19\%. Qwen3-0.6B-Non-Thinking exhibits a consistent trend: average accuracy improves by +3.01\% over OPD and surpasses FastOPD by +0.62\%. Relay-OPD further achieves significantly higher pass@$k$ than standard OPD across different sampling budgets (Figure~\ref{fig:hmmt-passk}).\par

\textbf{Comparison with trajectory intervention baselines.} TRD exhibits marked performance degradation relative to standard OPD on both student models: 30.69 vs.\ 41.23 on 1.7B and 18.07 vs.\ 28.03 on 0.6B. We observe that its rewritten trajectories often carry visible rewriting artifacts rather than resembling naturally unfolding problem-solving; examples appear in Appendix~\ref{app:trd-rewritten}. SKD only marginally outperforms OPD on 1.7B (42.35 vs.\ 41.23) and drops to 24.38 on 0.6B; it struggles to break established repetitive generation patterns (Appendix~\ref{app:skd-repetition}). FastOPD concentrates training signals at the sequence front via fixed truncation but cannot provide demonstrations of how to recover from failed prefixes. In contrast, Figure~\ref{fig:inference-length} shows that Relay-OPD reduces mean response length by 17.9\%, 14.2\%, and 28.3\% on AIME 2025, AIME 2026, and HMMT February 2026, respectively, compared to FastOPD, while simultaneously improving accuracy by +2.39\%, +4.17\%, and +1.14\%. Relay-OPD teaches the student to redirect reasoning before deviations compound further, yielding more accurate answers with shorter reasoning processes.\par

\textbf{Training token efficiency.} Relay-OPD uses shorter training trajectories and reaches the best checkpoint in fewer update steps. For the 1.7B student, Relay-OPD reaches its optimum at step 35, earlier than OPD at step 55 and FastOPD at step 45; its average rollout response length is 2,296 tokens, a 50.7\% reduction from OPD's 4,658 and shorter than FastOPD's 2,709. A consistent trend holds for the 0.6B student: Relay-OPD's average training trajectory length is 2,490 tokens, a 63.9\% reduction from OPD.\par

\subsection{Training Dynamics}\label{subsec:training-dynamics}

\begin{figure}[t]
\centering
\includegraphics[width=1.00\linewidth]{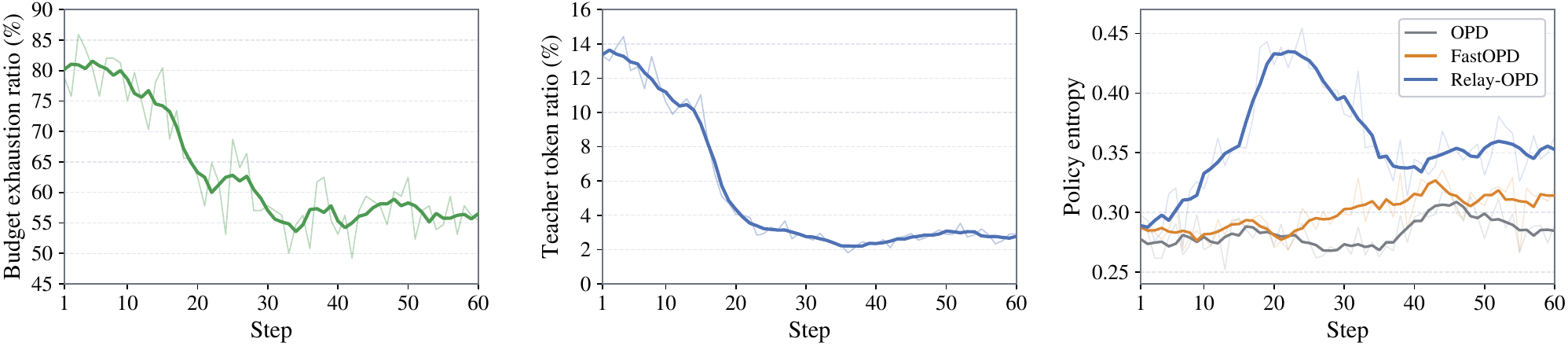}
\caption{\textbf{Training Dynamics of Relay-OPD.} Using Qwen3-4B-Instruct-2507 as teacher and Qwen3-1.7B-Non-Thinking as student. \textbf{Left:} fraction of trajectories exhausting the relay budget; \textbf{Middle:} teacher token ratio relative to all effective response tokens; \textbf{Right:} policy entropy of Relay-OPD, OPD, and FastOPD.}
\label{fig:training-dynamics}
\end{figure}

To elucidate how Relay-OPD adjusts teacher intervention as the student policy evolves, we track three metrics over the first 60 training steps in Figure~\ref{fig:training-dynamics}.
The left panel shows that the fraction of trajectories exhausting the relay budget gradually decreases, from roughly 75\%--85\% early on to roughly 50\%--60\%. The middle panel shows that teacher token ratio starts at roughly 13\%, then drops rapidly and stabilizes at 2\%--3\% after about 20 steps. Both trends indicate that the teacher--student gap and the prefix failures requiring intervention shrink during training.\par

The right panel compares policy entropy across the three methods. Relay-OPD maintains consistently higher entropy than OPD and FastOPD over the first 60 steps. We attribute this to teacher intervention altering the prefixes the student visits and the subsequent generation directions, thereby increasing student exploration.\par

\subsection{Ablation Studies}\label{subsec:ablation}

\textbf{Role of the teacher leg.} The benefit of the teacher leg stems from corrected context and local reasoning demonstrations, not merely early termination. To isolate this contribution, we set the maximum number of takeovers to $M=1$ for both variants: Trigger-stop terminates immediately at the first trigger without generating any teacher tokens; Relay-OPD generates a teacher leg of length $L=3$ before terminating.\par

As shown in Table~\ref{tab:teacher-leg-trigger-stop}, adding the teacher leg improves average accuracy from 43.48 to 46.25 (+2.77\% over Trigger-stop), confirming that corrected context and local reasoning demonstrations yield benefits beyond dynamic truncation alone.\par

\begin{table}[t]
\centering
\caption{\textbf{Teacher-Leg Ablations.} Using Qwen3-1.7B-Non-Thinking as student; all other settings are identical to the main experiments. \textbf{Top:} teacher leg vs.\ trigger-stop, where both variants set $M=1$. \textbf{Bottom:} training objective for the teacher leg.}
\label{tab:teacher-leg-trigger-stop}
\label{tab:teacher-leg-objective}
\begingroup
\small
\renewcommand{\arraystretch}{1.1}
\setlength{\tabcolsep}{3pt}
\resizebox{\linewidth}{!}{%
\begin{tabular}{lccccccccc}
\toprule
\textbf{Variant} & \textbf{AIME24} & \textbf{AIME25} & \textbf{AIME26} & \textbf{MATH} & \textbf{AMC23} & \textbf{Olymp.} & \textbf{\makecell{HMMT\\Feb26}} & \textbf{\makecell{HMMT\\Nov25}} & \textbf{Avg} \\
\midrule
\multicolumn{10}{c}{\textit{Teacher leg vs.\ trigger-stop}} \\
\midrule
Trigger-Stop ($M=1$, No Teacher Leg) & 39.48 & 28.75 & 24.48 & 87.00 & 71.56 & 55.27 & 23.11 & 18.23 & 43.48 \\
\rowcolor{cyan!10}
\textbf{Relay-OPD} ($M=1$, $L=3$) & \textbf{42.40} & \textbf{31.98} & \textbf{29.69} & \textbf{88.55} & \textbf{75.47} & \textbf{58.38} & \textbf{23.30} & \textbf{20.21} & \textbf{46.25} \\
\midrule
\multicolumn{10}{c}{\textit{Training objective for the teacher leg}} \\
\midrule
Student Draft Token & 38.54 & 27.60 & 26.25 & 88.10 & 75.94 & 57.42 & 23.67 & 18.96 & 44.56 \\
Teacher FKL ($k=128$) & 41.77 & 28.96 & 26.56 & 88.20 & 72.73 & 55.68 & 21.69 & 17.08 & 44.08 \\
\rowcolor{cyan!10}
Relay Token (Ours) & \textbf{42.71} & \textbf{32.81} & \textbf{30.52} & \textbf{89.50} & \textbf{76.88} & \textbf{58.79} & \textbf{24.72} & 19.79 & \textbf{46.96} \\
\bottomrule
\end{tabular}%
}
\endgroup
\end{table}

\textbf{Training objective for the teacher leg.} We compare two alternative objectives for the teacher leg, keeping the student leg under the standard OPD objective throughout. To support this ablation, the engine (\S\ref{subsec:implementation}) saves position markers for teacher legs together with the student's draft token $a_t^S$ on prefix $h_t^z$ at each teacher-leg position, before the teacher verifies or replaces it. The Student draft token variant uses the engine-saved $a_t^S$ and replaces the default advantage with\par

\begin{equation}
A_t^{\mathrm{draft}}
=
\log\pi_T(a_t^S\mid h_t^z)
-\log\pi_{\bar\theta}(a_t^S\mid h_t^z).
\end{equation}

That is, the policy gradient loss and parameter update at teacher-leg positions are computed with respect to the student's draft token $a_t^S$ rather than the actually generated relay token $z_t$.\par

The Teacher FKL variant takes the teacher's top-$k$ support set $\mathcal K_t=\operatorname{TopK}_k\!\left(\pi_T(\cdot\mid h_t^z)\right)$ with $k=128$. The teacher probability is renormalized over $\mathcal K_t$ as\par

\begin{equation}
q_T^k(v\mid h_t^z)
=
\frac{\pi_T(v\mid h_t^z)}
{\sum_{u\in\mathcal K_t}\pi_T(u\mid h_t^z)},
\qquad v\in\mathcal K_t.
\end{equation}

The corresponding Teacher FKL objective is\par

\begin{equation}
\mathcal L_{\mathrm{FKL}}^{\mathrm{teacher}}(t)
=
\sum_{v\in\mathcal K_t}
q_T^k(v\mid h_t^z)
\left[
\log q_T^k(v\mid h_t^z)
-\log\pi_\theta(v\mid h_t^z)
\right].
\end{equation}

As shown in Table~\ref{tab:teacher-leg-objective}, the relay token objective (46.96) substantially outperforms Student draft token (44.56) and Teacher FKL (44.08). The decline with Teacher FKL aligns perfectly with the design rationale in \S\ref{subsec:relay-opd}: the reverse-KL-style single-sample objective is inherently mode-seeking, enabling the student to selectively absorb the teacher's corrective signals on failed prefixes. In contrast, the mode-covering nature of Teacher FKL forces the student to match the teacher's full distribution, indiscriminately incorporating unreliable guidance. Student draft token predominantly suppresses tokens the student originally tended to generate, whereas using the actually generated tokens after takeover provides a more explicit learning signal.\par

\subsection{Sensitivity Analysis}\label{subsec:sensitivity}

\begin{figure}[t]
\centering
\includegraphics[width=1.00\linewidth]{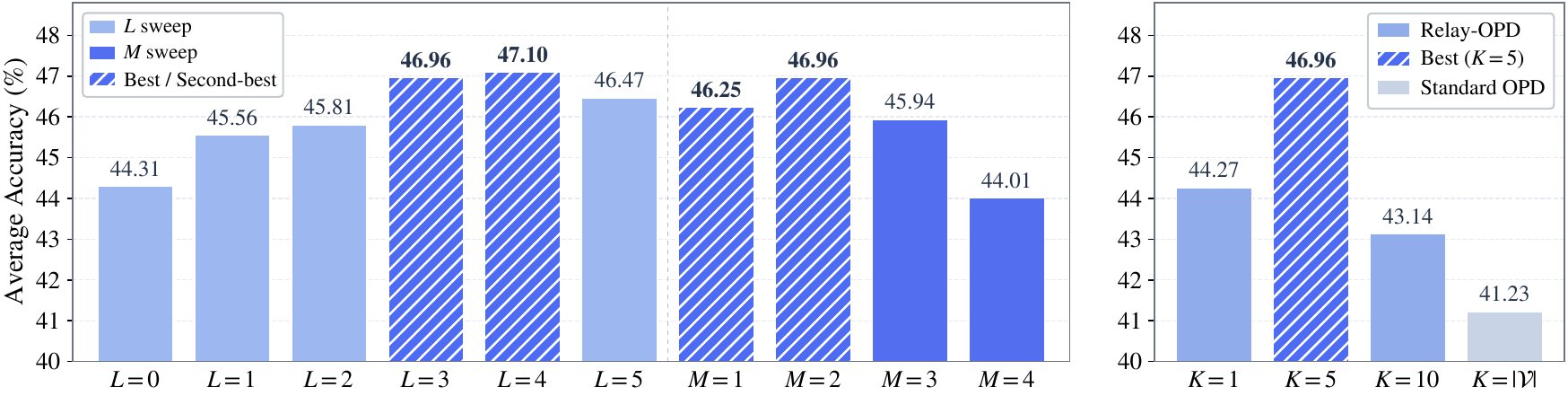}
\caption{\textbf{Hyperparameter Sensitivity of Relay-OPD.} Using Qwen3-4B-Instruct-2507 as teacher and Qwen3-1.7B-Non-Thinking as student. \textbf{Left:} relay budget $(M,L)$; \textbf{Right:} handoff top-$K$, where $|\mathcal V|$ denotes the vocabulary size.}
\label{fig:sensitivity-analysis}
\end{figure}

\textbf{Moderate intervention balances teacher correction against the on-policy property of the trajectory.} Figure~\ref{fig:sensitivity-analysis} summarizes average accuracy when varying one hyperparameter at a time; complete per-benchmark results appear in Appendix~\ref{app:sensitivity-full}.\par

\textbf{Relay budget $(M,L)$.} As shown in the left panel, average accuracy improves from 44.31 to 47.10 as $L$ increases from 0 to 4, then drops to 46.47 at $L=5$; both $L=3$ and $L=4$ achieve strong performance. For $M$, $M=1$ and $M=2$ achieve 46.25 and 46.96, whereas $M=4$ drops to 44.01. Figure~\ref{fig:motivation-intervention} already showed that extending teacher takeover yields diminishing returns; training results here further confirm that overly long or frequent teacher takeovers push the trajectory too far from the student's current policy, undermining the advantages of on-policy distillation.\par

\textbf{Handoff top-$K$.} As shown in the right panel, $K=1$, $K=5$, and $K=10$ achieve average accuracies of 44.27, 46.96, and 43.14---all exceeding the 41.23 of standard OPD (equivalent to $K=|\mathcal V|$), validating state-driven selective intervention. $K=5$ achieves the best result, indicating that the divergence threshold $K$ should be neither too small (over-triggering on minor differences) nor too large (missing genuine divergences).\par

\section{Conclusion}\label{sec:conclusion}

We introduced Relay-OPD, which addresses the prefix failure problem in on-policy distillation by detecting teacher--student reasoning-direction divergences online and letting the teacher intervene locally at detected failure points. A limited relay budget concentrates intervention on critical early positions while limiting departure from the student policy. Across two Qwen3 student models and eight mathematical reasoning benchmarks, Relay-OPD outperforms standard OPD by +5.73\% and the strongest trajectory intervention baseline by +1.49\% on average, while reducing training trajectory length by over 50\%. Further ablations and sensitivity analyses validate the contribution of each design component.\par

\bibliographystyle{plainnat}
\bibliography{references}

\clearpage
\appendix
\section*{Supplementary Material}
\subsection*{Contents}
\noindent\hyperref[app:training-config]{\ref*{app:training-config}\quad \nameref*{app:training-config}}\dotfill\pageref*{app:training-config}\par
\noindent\hyperref[app:prompts]{\ref*{app:prompts}\quad \nameref*{app:prompts}}\dotfill\pageref*{app:prompts}\par
\noindent\hyperref[app:algorithm-baselines]{\ref*{app:algorithm-baselines}\quad \nameref*{app:algorithm-baselines}}\dotfill\pageref*{app:algorithm-baselines}\par
\noindent\hyperref[app:supplementary]{\ref*{app:supplementary}\quad \nameref*{app:supplementary}}\dotfill\pageref*{app:supplementary}\par
\noindent\hyperref[app:case-studies]{\ref*{app:case-studies}\quad \nameref*{app:case-studies}}\dotfill\pageref*{app:case-studies}\par
\noindent\hyperref[app:limitations]{\ref*{app:limitations}\quad \nameref*{app:limitations}}\dotfill\pageref*{app:limitations}\par
\medskip

\needspace{8\baselineskip}\section{Training Configuration}\label{app:training-config}

\subsection{Handoff Trigger Word List}\label{app:trigger-words}

The reflection token set $\mathcal R$ used in the handoff criterion consists of the following base words together with their case and leading-space variants:\par

\begin{promptbox}[title={Handoff Trigger Word List}, breakable=false]
\ttfamily
Wait, But, Hmm, Actually, Hold, However, Yet, Oh, Alternatively, No, Ah, Oops, Well
\end{promptbox}

\subsection{Training Hyperparameters}\label{app:hyperparameters}

The complete training hyperparameters for the online distillation methods and for GRPO are listed in Table~\ref{tab:hyperparameters-online} and Table~\ref{tab:hyperparameters-grpo}, respectively.\par

\begin{table}[H]
\centering
\begingroup
\small
\renewcommand{\arraystretch}{1.1}
\setlength{\tabcolsep}{2pt}
\begin{minipage}[t]{0.485\linewidth}
\vspace{0pt}
\begin{minipage}[t][3.5\baselineskip][t]{\linewidth}
\caption{Training hyperparameters for online distillation methods.}
\label{tab:hyperparameters-online}
\end{minipage}
\begin{tabularx}{\linewidth}{L>{\raggedright\arraybackslash}p{0.27\linewidth}}
\toprule
\textbf{Hyperparameter} & \textbf{Value} \\
\midrule
Max Prompt Length & 2048 \\
Max Response Length & 16384 \\
Rollout Temperature & 1.0 \\
Rollout Top-$p$ & 1.0 \\
Rollout Number per Prompt ($n$) & 1 \\
Global Batch Size & 128 \\
PPO Mini-Batch Size & 128 \\
PPO Epochs & 1 \\
PPO Clipping Range & 0.2 \\
Learning Rate & $1\times10^{-6}$ \\
Learning-Rate Schedule & constant \\
Training Epochs & 1 \\
\bottomrule
\end{tabularx}
\end{minipage}\hfill
\begin{minipage}[t]{0.485\linewidth}
\vspace{0pt}
\begin{minipage}[t][3.5\baselineskip][t]{\linewidth}
\caption{Training hyperparameters for GRPO.}
\label{tab:hyperparameters-grpo}
\end{minipage}
\begin{tabularx}{\linewidth}{L>{\raggedright\arraybackslash}p{0.27\linewidth}}
\toprule
\textbf{Hyperparameter} & \textbf{Value} \\
\midrule
Max Prompt Length & 2048 \\
Max Response Length & 16384 \\
Rollout Temperature & 1.0 \\
Rollout Top-$p$ & 1.0 \\
Rollout Number per Prompt ($n$) & 8 \\
Global Batch Size & 128 \\
PPO Mini-Batch Size & 128 \\
PPO Epochs & 1 \\
PPO Clipping Range & 0.2 \\
Learning Rate & $1\times10^{-6}$ \\
Learning-Rate Schedule & constant \\
Training Epochs & 1 \\
\bottomrule
\end{tabularx}
\end{minipage}
\endgroup
\end{table}

\needspace{8\baselineskip}\section{Prompts}\label{app:prompts}

\subsection{Training and Inference Prompt}\label{app:training-prompt}

Both training and inference use the following template:\par

\begin{promptbox}[title={Training and Inference Prompt}, breakable=false]
$<|$im\_start$|>$system\\
Please reason step by step, and put your final answer within \textbackslash boxed\{\}.$<|$im\_end$|>$\\
$<|$im\_start$|>$user\\
\{problem\}$<|$im\_end$|>$\\
$<|$im\_start$|>$assistant\\
$<$think$>$\\[\baselineskip]
$<$/think$>$
\end{promptbox}

\subsection{TRD Rewrite Prompt}\label{app:trd-prompt}

We adopt the rewrite prompt from the original TRD paper \citep{jiang2026trajectory}, with user content as follows:\par

\begin{promptbox}[title={TRD Rewrite Prompt}, breakable=false]
Your task is to rewrite your mathematical solution.\\[\baselineskip]
**Problem:**\\
\{problem\}\\[\baselineskip]
**Your Initial Solution:**\\
\{initial\_response\}\\[\baselineskip]
**Instructions:**\\
1. Preserve the overall structure and reasoning path of your original solution\\
2. Identify and fix errors in computation or logic\\
3. Keep correct intermediate steps and meaningful work\\
4. Output ONLY the rewritten solution
\end{promptbox}

\needspace{8\baselineskip}\section{Algorithm and Baselines}\label{app:algorithm-baselines}

\subsection{Relay-OPD Algorithm}\label{app:relay-algorithm}

The relay rollout construction procedure of Relay-OPD is given in Algorithm~\ref{alg:relay-rollout}.\par

\begin{algorithm}[H]
\caption{Relay Rollout Construction}
\label{alg:relay-rollout}
\begin{algorithmic}[1]
\Require prompt $x$, student policy $\pi_{\bar\theta}$, teacher policy $\pi_T$, reflection-token set $\mathcal R$, handoff top-$K$, relay budget $(M,L)$
\Ensure relay trajectory $z$
\State Initialize $z\leftarrow()$, takeover count $j\leftarrow0$, state $s\leftarrow\mathsf{S}$
\While{neither EOS nor the maximum length is reached}\Comment{either event enters the terminal state $\bot$}
    \State \textcolor{green!55!black}{// Student drafting and handoff detection}
    \State Set prefix $h\leftarrow(x,z)$ and sample student draft token $a^S\sim\pi_{\bar\theta}(\cdot\mid h)$
    \State Compute teacher argmax $a^T\leftarrow\arg\max_{v}\pi_T(v\mid h)$ and student top-$K$ set $\mathcal K_S(h)\leftarrow\operatorname{TopK}_K(\pi_{\bar\theta}(\cdot\mid h))$
    \State Evaluate the handoff criterion $\phi(h)\leftarrow\mathbf 1[a^T\in\mathcal R]\cdot\mathbf 1[\mathcal K_S(h)\cap\mathcal R=\varnothing]$
    \If{$s=\mathsf{S}$ \textbf{and} $\phi(h)=1$ \textbf{and} $j<M$}\Comment{$\mathsf{S}\!\to\!\mathsf{T}$ transition}
        \State \textcolor{green!55!black}{// Teacher leg}
        \State $j\leftarrow j+1$, $s\leftarrow\mathsf{T}$; append $a^T$ to $z$\Comment{reflection token opens the teacher leg}
        \For{$\ell=1,\ldots,L$}
            \State Generate one teacher paragraph via speculative decoding (\S\ref{subsec:implementation}) and append it to $z$
        \EndFor
        \If{$j=M$}
            \State \textbf{break}\Comment{relay budget exhausted; enter the terminal state $\bot$}
        \Else
            \State $s\leftarrow\mathsf{S}$\Comment{$\mathsf{T}\!\to\!\mathsf{S}$ transition: the student resumes}
        \EndIf
    \Else
        \State Append $a^S$ to $z$\Comment{student leg, including $\phi(h)=1$ with $j=M$}
    \EndIf
\EndWhile
\end{algorithmic}
\end{algorithm}

\subsection{Method Comparison Overview}\label{app:method-comparison}

We summarize all compared methods in Table~\ref{tab:method-comparison} along policy type, trajectory source, training loss, and method-specific settings.\par

\begin{table}[htb]
\centering
\caption{\textbf{Method Comparison Overview.} Original FKL-based papers use full vocabulary or varying top-$k$ settings; we uniformly adopt top-$k{=}128$. The original SKD paper uses acceptance top-$k{=}25$, but in our teacher--student configuration top-25 barely filters any student draft tokens, so we adopt top-$k{=}5$.}
\label{tab:method-comparison}
\begingroup
\small
\renewcommand{\arraystretch}{1.1}
\setlength{\tabcolsep}{3pt}
\renewcommand{\tabularxcolumn}[1]{m{#1}}
\begin{tabularx}{\linewidth}{>{\raggedright\arraybackslash}m{0.14\linewidth}>{\centering\arraybackslash}m{0.13\linewidth}>{\raggedright\arraybackslash}m{0.20\linewidth}>{\centering\arraybackslash}m{0.09\linewidth}L}
\toprule
\textbf{Method} & \textbf{Policy} & \textbf{Trajectory Source} & \textbf{Loss} & \textbf{Method-Specific Settings} \\
\midrule
SFT & Off-policy & Teacher trajectory & CE & — \\
KD & Off-policy & Teacher trajectory & FKL & — \\
TRD & Off-policy & Rewritten student trajectory & FKL & rewrite prompt (B.2) \\
GRPO & On-policy & Student rollout & Outcome RL & 8 rollouts per prompt \\
OPD & On-policy & Student rollout & RKL & max response = 16384 \\
FastOPD & On-policy & Student rollout & RKL & fixed truncation $\in\{1024,2048,4096,8192\}$ (C.5) \\
SKD & On-policy & Speculative mixed rollout & FKL & acceptance top-$K{=}5$ (C.4) \\
Relay-OPD & On-policy & Relay rollout & RKL & handoff top-$K$; relay budget $(M,L)$ \\
\bottomrule
\end{tabularx}
\endgroup
\end{table}

\subsection{TRD}\label{app:trd}

TRD \citep{jiang2026trajectory} constructs training data offline in two stages. Stage one lets the student generate an initial trajectory $y_o$ from the original problem $x$; stage two provides $x$ and $y_o$ to the teacher within a rewrite prompt, producing a rewritten trajectory $y_r$. We adopt the rewrite prompt from the original TRD paper; the complete prompt is given in Appendix~\ref{app:trd-prompt}.\par

During TRD training, teacher and student use different contexts. For the $t$-th token in the rewritten trajectory, the student context is $h_t^S=(x,y_{r,<t})$, while the teacher context additionally includes the student's initial trajectory: $h_t^T=(x,y_o,y_{r,<t})$. On teacher context $h_t^T$, the top-$k$ support set is $\mathcal K_t^T:=\operatorname{TopK}_k\!\left(\pi_T(\cdot\mid h_t^T)\right)$ with $k=128$, and teacher probability is renormalized over this support set as\par

\begin{equation}
q_T^k(v\mid h_t^T)
:=
\frac{\pi_T(v\mid h_t^T)}
{\sum_{u\in\mathcal K_t^T}\pi_T(u\mid h_t^T)},
\qquad v\in\mathcal K_t^T.
\end{equation}

The corresponding top-$k$ FKL objective is\par

\begin{equation}
\mathcal L_{\mathrm{TRD}}(t)
=
\sum_{v\in\mathcal K_t^T}
q_T^k(v\mid h_t^T)
\left[
\log q_T^k(v\mid h_t^T)
-\log\pi_\theta(v\mid h_t^S)
\right].
\end{equation}

This asymmetry means that the teacher provides supervision conditioned on $y_o$, while the student cannot access $y_o$ at either training or inference time.\par

\subsection{SKD}\label{app:skd}

SKD \citep{xu2025speculative} constructs teacher--student mixed trajectories via speculative decoding. The student serves as draft model and the teacher as target model; throughout this subsection, $z$ denotes SKD's mixed trajectory and $h_t=(x,z_{<t})$ its prefix. Given prefix $h_t$ and student draft token $a_t^S$, the teacher's top-$K$ support set is $\mathcal K_T(h_t)=\operatorname{TopK}_K\!\left(\pi_T(\cdot\mid h_t)\right)$ with $K=5$, and the draft is accepted iff $a_t^S\in\mathcal K_T(h_t)$; if accepted, $z_t=a_t^S$. If rejected, a replacement token is sampled from the teacher's distribution renormalized over $\mathcal K_T(h_t)\setminus\{a_t^S\}$:\par

\begin{equation}
z_t
\sim
\widetilde\pi_T(\cdot\mid h_t),
\qquad
\operatorname{supp}(\widetilde\pi_T)=\mathcal K_T(h_t)\setminus\{a_t^S\}.
\end{equation}

Replaced positions are marked as teacher-owned. The training objective uses top-$k$ FKL on the mixed trajectory, with teacher and student sharing the same prefix $h_t$.\par

\subsection{FastOPD}\label{app:fastopd}

FastOPD \citep{ziheng2026less,zhang2026fast} truncates the student rollout at a fixed length. Given student-generated trajectory $y=(y_1,\ldots,y_N)$ and truncation length $B$, the training trajectory is $y^{(B)}=(y_1,\ldots,y_{N_B})$ with $N_B=\min(N,B)$.\par

FastOPD then optimizes the standard OPD objective (\S\ref{subsec:opd}) on $y^{(B)}$, replacing $N$ and $y$ with $N_B$ and $y^{(B)}$. Thus FastOPD changes only the rollout length participating in training, not the token-level learning signal of standard OPD.\par

\clearpage\section{Supplementary Results}\label{app:supplementary}

\subsection{Complete Sensitivity Results}\label{app:sensitivity-full}

Complete per-benchmark results for the relay budget and handoff top-$K$ sweeps of \S\ref{subsec:sensitivity} are reported in Table~\ref{tab:sensitivity-results}.\par

\begin{table}[H]
\centering
\caption{\textbf{Complete Sensitivity Results.} Using Qwen3-4B-Instruct-2507 as teacher and Qwen3-1.7B-Non-Thinking as student, varying one hyperparameter at a time. Standard OPD corresponds to $K=|\mathcal{V}|$; bold denotes the best result in each column.}
\label{tab:sensitivity-results}
\begingroup
\small
\renewcommand{\arraystretch}{1.1}
\setlength{\tabcolsep}{3pt}
\resizebox{\linewidth}{!}{%
\begin{tabular}{lccccccccc}
\toprule
\textbf{Variant} & \textbf{AIME24} & \textbf{AIME25} & \textbf{AIME26} & \textbf{MATH} & \textbf{AMC23} & \textbf{Olymp.} & \textbf{\makecell{HMMT\\Feb26}} & \textbf{\makecell{HMMT\\Nov25}} & \textbf{Avg} \\
\midrule
$L=0$ & 40.42 & 29.06 & 27.19 & 88.15 & 73.83 & 56.31 & 23.48 & 16.04 & 44.31 \\
$L=1$ & 41.46 & 31.46 & 26.88 & 88.75 & 76.64 & 57.68 & 22.73 & 18.85 & 45.56 \\
$L=2$ & 40.52 & 32.50 & 28.23 & 88.55 & 78.20 & 57.23 & 22.16 & 19.06 & 45.81 \\
$L=4$ & 42.60 & 32.40 & \textbf{32.71} & \textbf{89.50} & 78.20 & 58.49 & 23.39 & 19.48 & \textbf{47.10} \\
$L=5$ & 40.00 & \textbf{32.81} & 30.10 & 89.35 & \textbf{78.28} & 57.94 & 23.58 & 19.69 & 46.47 \\
$M=1$ & 42.40 & 31.98 & 29.69 & 88.55 & 75.47 & 58.38 & 23.30 & 20.21 & 46.25 \\
$M=3$ & \textbf{43.12} & 31.46 & 26.88 & 88.85 & 76.48 & 57.46 & 22.82 & \textbf{20.42} & 45.94 \\
$M=4$ & 37.60 & 32.08 & 26.88 & 86.90 & 73.52 & 56.34 & 21.59 & 17.19 & 44.01 \\
$K=1$ & 40.00 & 29.90 & 24.38 & 87.55 & 74.22 & 56.86 & 22.73 & 18.54 & 44.27 \\
$K=10$ & 36.46 & 29.27 & 26.04 & 88.15 & 73.05 & 56.19 & 20.36 & 15.62 & 43.14 \\
$K=\lvert\mathcal{V}\rvert$ & 35.83 & 25.52 & 23.33 & 85.70 & 70.08 & 55.27 & 20.08 & 14.06 & 41.23 \\
\makecell[l]{\textbf{Default}\\($K=5$, $M=2$, $L=3$)} & 42.71 & \textbf{32.81} & 30.52 & \textbf{89.50} & 76.88 & \textbf{58.79} & \textbf{24.72} & 19.79 & 46.96 \\
\bottomrule
\end{tabular}%
}
\endgroup
\end{table}

\subsection{FastOPD Truncation Length Sweep}\label{app:fastopd-sweep}

Complete results across fixed truncation lengths are reported in Table~\ref{tab:fastopd-truncation}. Both student models achieve the highest average accuracy at 4,096 tokens; Table~\ref{tab:main-results} therefore reports this configuration.\par

\begin{table}[H]
\centering
\caption{\textbf{FastOPD Fixed Truncation Length Comparison.} Subscripts in the Avg column indicate the training step of the best checkpoint; Train Len denotes average rollout response length from training start to that checkpoint.}
\label{tab:fastopd-truncation}
\begingroup
\small
\renewcommand{\arraystretch}{1.1}
\setlength{\tabcolsep}{3pt}
\resizebox{\linewidth}{!}{%
\begin{tabular}{ccccccccccc}
\toprule
\textbf{\makecell{Truncation\\Length}} & \textbf{AIME24} & \textbf{AIME25} & \textbf{AIME26} & \textbf{MATH} & \textbf{AMC23} & \textbf{Olymp.} & \textbf{\makecell{HMMT\\Feb26}} & \textbf{\makecell{HMMT\\Nov25}} & \textbf{Avg} & \textbf{\makecell{Train\\Len}} \\
\midrule
\multicolumn{11}{c}{\textit{Student: Qwen3-0.6B-Non-Thinking}} \\
\midrule
1,024 & 15.94 & 20.42 & 12.40 & 74.65 & 54.69 & 43.95 & 10.51 & 9.58 & 30.27$_{\text{@105}}$ & 987 \\
2,048 & 15.83 & 19.38 & 15.42 & 76.90 & 51.48 & 44.77 & 10.51 & 8.33 & 30.33$_{\text{@95}}$ & 1,827 \\
\textbf{4,096} & 15.83 & 20.10 & 13.54 & 75.30 & 53.67 & 44.81 & 11.55 & 8.54 & \textbf{30.42}$_{\text{@100}}$ & 3,302 \\
8,192 & 14.79 & 18.44 & 11.88 & 75.35 & 51.80 & 43.36 & 9.66 & 8.65 & 29.24$_{\text{@60}}$ & 5,055 \\
\midrule
\multicolumn{11}{c}{\textit{Student: Qwen3-1.7B-Non-Thinking}} \\
\midrule
1,024 & 39.06 & 26.46 & 25.21 & 86.00 & 70.70 & 57.08 & 23.20 & 19.17 & 43.36$_{\text{@105}}$ & 983 \\
2,048 & 43.23 & 29.58 & 28.33 & 88.35 & 72.11 & 57.42 & 23.67 & 19.17 & 45.23$_{\text{@75}}$ & 1,753 \\
\textbf{4,096} & 42.29 & 30.42 & 26.35 & 87.95 & 74.30 & 58.16 & 23.58 & 20.73 & \textbf{45.47}$_{\text{@45}}$ & 2,709 \\
8,192 & 40.52 & 27.60 & 26.56 & 86.70 & 72.81 & 56.27 & 21.88 & 20.31 & 44.08$_{\text{@50}}$ & 4,004 \\
\bottomrule
\end{tabular}%
}
\endgroup
\end{table}

\clearpage\section{Case Studies}\label{app:case-studies}

\subsection{Relay-OPD Takeover Case}\label{app:takeover-case}

In this case, the student has only 5 RM remaining after purchasing 15 apples yet incorrectly concludes it can still afford both a mango and a papaya. Without teacher intervention, it rationalizes this plan into a total of 17 fruits. Faced with the same prefix, the teacher selects \texttt{But} with 74.4084\% probability (Table~\ref{tab:teacher-student-distributions}), whereas the student favors \texttt{So} or \texttt{Now} to extend the current conclusion. The teacher leg points out that the two fruits would cost at least 9 RM, and the student accordingly rejects the original plan and finds the feasible combination of 12 apples, 1 mango, and 2 papayas, arriving at the correct answer of $15$. This case illustrates two points: the handoff trigger correctly identifies the moment the student would continue along a wrong direction, and a brief teacher takeover suffices to correct the subsequent reasoning.\par

\begin{promptbox}[title={Problem}, breakable=false]
\small
Danial went to a fruit stall that sells apples, mangoes, and papayas. Each apple costs $3$ RM, each mango costs $4$ RM, and each papaya costs $5$ RM. He bought at least one of each fruit and paid exactly $50$ RM. What is the maximum number of fruits that he could have bought?\par
\smallskip
Ground truth: \texttt{15}
\end{promptbox}

\begin{promptbox}[title={Student Trajectory (w/o Intervention)}, breakable=false]
\small
\ldots\par

Now, try \textbf{15 apples}:\par

\begin{itemize}[leftmargin=*,itemsep=0.25ex,topsep=0.5ex]
\item $15\times3=45$
\item Remaining: $50-45=5$
\end{itemize}

We can buy \textbf{1 papaya} (5 RM), and at least 1 mango (4 RM). \textbf{So that's valid.}\par

\textcolor{handoffred}{\textbf{[Handoff trigger detected; the following is the student's original continuation]}}\par

So:\par

\begin{itemize}[leftmargin=*,itemsep=0.25ex,topsep=0.5ex]
\item 15 apples
\item 1 mango
\item 1 papaya
\end{itemize}

Total fruits: $15+1+1=17$\par

Is this the \textbf{maximum} possible? Let's check.\par

Try \textbf{14 apples}: $14\times3=42$, remaining $50-42=8$; we can buy 1 mango (4 RM) and 1 papaya (5 RM). Total: $14+1+1=16$ fruits. Smaller than 17.\par

\ldots\par

Final Answer:\par

\begin{equation*}
\boxed{17}
\end{equation*}
\end{promptbox}

\begin{table}[H]
\centering
\caption{Teacher and student distributions.}
\label{tab:teacher-student-distributions}
\begingroup
\small
\renewcommand{\arraystretch}{1.1}
\setlength{\tabcolsep}{3pt}
\begin{tabularx}{\linewidth}{>{\centering\arraybackslash}p{0.10\linewidth}*{2}{L}}
\toprule
\textbf{Rank} & \textbf{Teacher distribution} & \textbf{Student distribution} \\
\midrule
1 & \texttt{But} 74.4084\% & \texttt{So} 50.6288\% \\
2 & \texttt{So} 10.0701\% & \texttt{Now} 21.1052\% \\
3 & \texttt{Wait} 7.8426\% & \texttt{-} 16.4368\% \\
4 & \texttt{We} 4.1978\% & \texttt{Total} 5.3362\% \\
5 & \texttt{-} 1.2027\% & \texttt{Then} 2.8563\% \\
6 & \texttt{Can} 0.5014\% & \texttt{That} 0.5624\% \\
7 & \texttt{Now} 0.2368\% & \texttt{F} 0.4963\% \\
8 & \texttt{Remaining} 0.2368\% & \texttt{This} 0.4380\% \\
9 & \texttt{Then} 0.2090\% & \texttt{We} 0.4380\% \\
10 & \texttt{Try} 0.1268\% & \texttt{Let} 0.3866\% \\
\bottomrule
\end{tabularx}
\endgroup
\end{table}

\begin{promptbox}[title={Relay Trajectory}, breakable=false]
\small
\ldots\par

Now, try \textbf{15 apples}:\par

\begin{itemize}[leftmargin=*,itemsep=0.25ex,topsep=0.5ex]
\item $15\times3=45$
\item Remaining: $50-45=5$
\end{itemize}

We can buy \textbf{1 papaya} (5 RM), and at least 1 mango (4 RM). \textbf{So that's valid.}\par

\textcolor{handoffred}{\textbf{[Handoff trigger detected; the teacher takes over]}}\par

\textbf{But we have only 5 RM left after 15 apples --- we would have to spend 4 RM on mango and 5 on papaya --- total 9, which exceeds 5.}\par

\textbf{Wait --- no: we have 5 RM left, and we must spend at least 4 RM for mango and 5 RM for papaya, but only 5 RM is available. So both cannot be bought.}\par

\textbf{We need at least one mango and one papaya.}\par

\textcolor{handoffred}{\textbf{[Teacher leg ends; generation returns to the student]}}\par

\textbf{So 15 apples is not valid.}\par

\ldots\par

Try \textbf{12 apples}:\par

\begin{itemize}[leftmargin=*,itemsep=0.25ex,topsep=0.5ex]
\item $12\times3=36$
\item Remaining: $50-36=14$
\end{itemize}

Try:\par

\begin{itemize}[leftmargin=*,itemsep=0.25ex,topsep=0.5ex]
\item 1 mango = 4, 2 papayas = 10 $\to$ total = 14 $\to$ \textbf{valid!}
\end{itemize}

So:\par

\begin{itemize}[leftmargin=*,itemsep=0.25ex,topsep=0.5ex]
\item 12 apples
\item 1 mango
\item 2 papayas
\end{itemize}

Total = $12+1+2=\mathbf{15}$ fruits\par

\ldots\par

\begin{equation*}
\boxed{15}
\end{equation*}
\end{promptbox}

\needspace{17\baselineskip}
\subsection{TRD Rewritten Text}\label{app:trd-rewritten}

We matched references to the original solution (e.g., \texttt{initial/original solution}), direct descriptions of the rewriting task (e.g., \texttt{rewrite/revision}), and reviewer-style openings (e.g., \texttt{after review/reevaluation}); 18.96\% of trajectories contain at least one such expression, confirming that these rewriting artifacts are not isolated. The following representative fragments are extracted directly from teacher-generated rewritten trajectories; ellipses indicate truncated content.\par

\begin{promptbox}[title={TRD Rewritten Fragments}, breakable=false]
\small
Given the complexity, and that \textbf{the original solution incorrectly assumed a simple parity, we must revise.} \ldots\ a common variant of this game has the property that the first player wins iff $(N\bmod 3=1)$. Even though our manual simulation showed $(N=6)$ \ldots\ \textbf{perhaps we made a mistake.}\par
\smallskip
\textbf{This may be incorrect based on simulation;} however, without complete DP simulation \ldots\ \textbf{this is the most reasonable periodic answer.} \ldots\ \textbf{we go with $\boxed{674}$.}\par
\end{promptbox}

\subsection{SKD Repetition Patterns}\label{app:skd-repetition}

SKD accepts a student draft token whenever it falls within the teacher's top-$K$ support set (Appendix~\ref{app:skd}). Because acceptance is driven by this generic distributional agreement, SKD struggles to break a repetitive generation pattern once the student has established it: the tokens that continue the repetition typically remain inside the teacher's top-$K$ candidates, even when the teacher itself would terminate or redirect. The following fragment from an SKD mixed trajectory is representative: the student repeatedly emits a completed final-answer block.\par

\begin{promptbox}[title={SKD Mixed Trajectory}, breakable=false]
\small
\ldots\par

Final Answer:\par

\begin{equation*}
\boxed{\text{No integer } n}
\end{equation*}

Final Answer:\par

\begin{equation*}
\boxed{\text{No integer } n}
\end{equation*}

\ldots\par
\end{promptbox}

\begin{table}[H]
\centering
\caption{Teacher and student distributions at the end of a repeated answer block.}
\label{tab:skd-repetition-distributions}
\begingroup
\small
\renewcommand{\arraystretch}{1.1}
\setlength{\tabcolsep}{3pt}
\begin{tabularx}{\linewidth}{>{\centering\arraybackslash}p{0.10\linewidth}*{2}{L}}
\toprule
\textbf{Rank} & \textbf{Teacher distribution} & \textbf{Student distribution} \\
\midrule
1 & \texttt{<|im\_end|>} 73.7035\% & \texttt{\textbackslash n\textbackslash n} 78.5176\% \\
2 & \texttt{\textbackslash n\textbackslash n} 16.4455\% & \texttt{<|im\_end|>} 19.8524\% \\
3 & \texttt{\.G\^a\v{l}} 6.8555\% & \texttt{\textbackslash n} 1.2691\% \\
4 & \texttt{~~\textbackslash n} 0.7226\% & \texttt{~or} 0.1516\% \\
5 & \texttt{~**} 0.5627\% & \texttt{\.G\^a\v{l}} 0.0558\% \\
\bottomrule
\end{tabularx}
\endgroup
\end{table}

Table~\ref{tab:skd-repetition-distributions} shows the teacher and student next-token distributions on the prefix ending at a completed answer block.\footnote{Token strings appear in the tokenizer's byte-level representation, which maps every byte to a printable character: \texttt{\.G} denotes a leading space, and \texttt{\^a\v{l}} are the first two UTF-8 bytes of a check-mark symbol (e.g., \checkmark), whose final byte would be completed by a subsequent token.} The teacher intends to terminate generation: its top-$1$ token is the end-of-sequence token \texttt{<|im\_end|>} with probability 73.7035\%. The student instead prefers to open yet another paragraph, placing 78.5176\% on \texttt{\textbackslash n\textbackslash n}. Under the acceptance criterion $a_t^S\in\mathcal K_T(h_t)$ with $K=5$, however, \texttt{\textbackslash n\textbackslash n} still ranks second in the teacher's support set (16.4455\%), so the student draft is accepted and the repetition continues---and because the same distributions recur on the extended prefix, the trajectory can degenerate into an \textbf{infinite loop} of final-answer blocks. Acceptance driven by generic distributional agreement thus provides no explicit signal that the established generation pattern itself has gone wrong, which underlies the repetition-related degradation of SKD discussed in \S\ref{subsec:main-results}.\par

\needspace{8\baselineskip}\section{Limitations}\label{app:limitations}

Our evaluation focuses on mathematical reasoning with Qwen3 teacher--student pairs, the strong-to-weak setting targeted by prior on-policy distillation work. The relay mechanism itself is task-agnostic, and we leave its application to other domains, such as code generation or agentic tool use, to future work; the reflection-token set in Appendix~\ref{app:trigger-words} may need adjustment when switching to a different model family. Relay-OPD also presumes a teacher whose continuations redirect failed prefixes more reliably than the student's, so its benefit is expected to diminish as the teacher--student capability gap narrows. Finally, the relay budget was tuned on the 1.7B student and reused unchanged for the 0.6B student; the sensitivity analysis in \S\ref{subsec:sensitivity} indicates robustness across moderate budget ranges, though new model pairs may benefit from re-tuning.\par

\end{document}